\documentclass[letterpaper]{article} 
\usepackage[]{aaai24}  
\usepackage{times}  
\usepackage{helvet}  
\usepackage{courier}  
\usepackage[hyphens]{url}  
\usepackage{graphicx} 
\usepackage{natbib}  
\usepackage{caption} 
\usepackage{algorithm}
\usepackage{algorithmic}

\usepackage{newfloat}
\usepackage{listings}
\usepackage{lipsum}

\usepackage{graphicx}
\usepackage{bm}
\usepackage{amsmath}
\usepackage{amsfonts}

\usepackage{helvet}
\usepackage{courier}
\usepackage{multirow}
\usepackage{bm}
\usepackage{amsmath}
\usepackage{mdwlist}

\usepackage{relsize}
\usepackage{amsfonts}
\usepackage{url}
\usepackage{graphicx}
\usepackage{subfigure}
\usepackage{caption}
\usepackage{hhline}
\usepackage{color}
\usepackage{colortbl}
\usepackage{booktabs}
\usepackage{tabularx}
\usepackage{amsmath}
\usepackage{makecell}

\DeclareCaptionStyle{ruled}{labelfont=normalfont,labelsep=colon,strut=off} 
\floatstyle{ruled}
\newfloat{listing}{tb}{lst}{}
\floatname{listing}{Listing}
\title{LLaMA Beyond English: An Empirical Study on Language Capability Transfer}
\author{
    Jun Zhao\equalcontrib,
    Zhihao Zhang\equalcontrib,
    Luhui Gao,
    Qi Zhang\thanks{Corresponding Author},
    Tao Gui,
    Xuanjing Huang
}
\affiliations{
    \textsuperscript{\rm 1}School of Computer Science, Fudan University\\

    \{zhaoj19,zhangzhihao19,qz,tgui\}@fudan.edu.cn\\
}

\usepackage{bibentry}

\begin{document}

\maketitle

\begin{abstract}
In recent times, substantial advancements have been witnessed in large language models (LLMs), exemplified by ChatGPT, showcasing remarkable proficiency across a range of complex tasks. However, many mainstream LLMs (e.g. LLaMA) are pretrained on English-dominant corpus, which limits their performance in other non-English languages. In this paper, we focus on how to effectively transfer the capabilities of language generation and following instructions to a non-English language.
To answer this question, we conduct an extensive empirical investigation based on LLaMA, accumulating over 1440 GPU hours. We analyze the impact of key factors such as vocabulary extension, further pretraining, and instruction tuning on transfer. To accurately assess the model's level of knowledge, we employ four widely used standardized testing benchmarks: C-Eval, MMLU, AGI-Eval, and GAOKAO-Bench. Furthermore, a comprehensive evaluation of the model's response quality is conducted, considering aspects such as accuracy, fluency, informativeness, logical coherence, and harmlessness, based on LLM-Eval, a benchmarks consisting instruction tasks from 17 diverse categories.
Our evaluation results demonstrate that comparable performance to state-of-the-art transfer models can be achieved with less than $1\%$ of the pretraining data, both in terms of knowledge alignment and response quality. Furthermore, the experimental outcomes across the thirteen low-resource languages also exhibit similar trends. We anticipate that the conclusions revealed by the experiments will aid the community in developing non-English LLMs.

\end{abstract}

\section{Introduction}
    \begin{figure}[t]
        \includegraphics[width=\columnwidth]{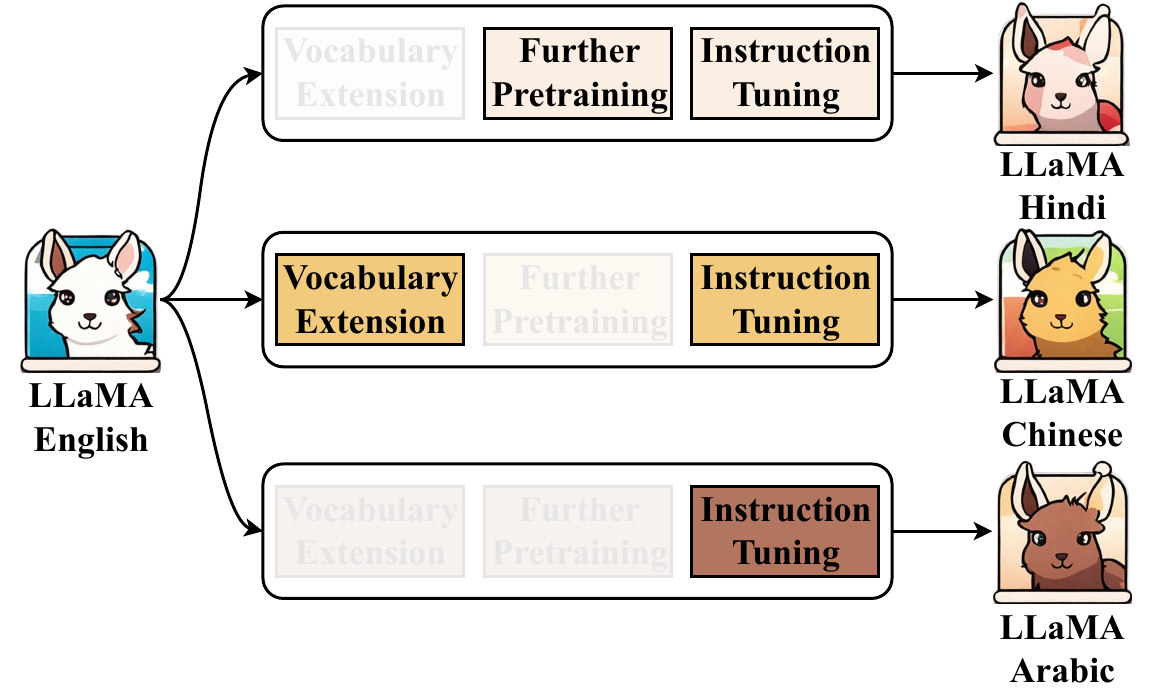}
        \caption{Pretrained LLaMA models, which are primarily trained on English-dominated corpus (as depicted on the left), are not inherently proficient in handling non-English languages. 
        We aim to investigate the necessity of vocabulary extension, further pretraining, and instruction tuning, as well as to what extent they influence the capability transfer.
        This exploration enables us to efficiently transfer LLaMA's language capabilities to non-English languages (as illustrated on the right), minimizing costs in the process.}
        \label{fig:intro}
    \end{figure}
For decades, researchers in Natural Language Processing (NLP) have been exploring the fundamental principles of intelligence \cite{bubeck2023sparks}. The recent advances in large language models (LLMs) seem to have revealed a glimmer of hope. Benefitting from the unprecedented scales of model size and training data, many LLMs like ChatGPT \cite{chatgpt}, PaLM \cite{anil2023palm}, LLaMA \cite{touvron2023llama}, and others have emerged strong capabilities in reasoning \cite{DBLP:journals/corr/abs-2110-14168}, planning \cite{pmlr-v162-huang22a}, and learning from experience \cite{dong2023survey} at or surpassing human levels. These general capabilities also provide a foundation for LLMs to address intricate real-world tasks, such as successfully completing the entire Uniform Bar Examination (UBE) \cite{katz2023gpt} or coding based on natural language instructions \cite{StableCode}.

Many well-known LLMs are capable of comprehending input and generating responses across different languages, thanks to their pretraining on a diverse mix of corpus from multiple languages. However, due to the imbalanced distribution of language resources, collecting extensive training data for all languages is nearly impossible \cite{ranta2021linguistic}. Taking the representative LLM BLOOM \cite{workshop2023bloom} as an example, it has been pretrained on 46 natural languages. Yet, this number accounts for only $0.66\%$ of the roughly $7,000$ languages currently in use. Moreover, within the corpus of these 46 languages, there exists extreme imbalance, with the high-resource English texts being 2.8 million times more than that of the low-resource Chitumbuka language. This is not an isolated case. Another widely discussed language model, LLaMA, has been pretrained primarily on English-dominated corpus, supplemented with limited data from 20 related languages that utilize the Latin and Cyrillic scripts. As a result, LLaMA exhibits inferior performance in contexts involving non-English languages where it has not undergone sufficient training. Some researchers collect large-scale data for specific languages of interest and retrain an LLM \cite{team2023internlm}. However, this inevitably leads to high computational and data collection costs, which is not suitable for low-resource languages. While \citet{cui2023efficient} extend original vocabulary and further pretrain LLaMA with 30B Chinese tokens by LoRA \cite{DBLP:journals/corr/abs-2106-09685}, reporting promising results. Nonetheless, a fine-grained systematic investigation of the transfer process remains lacking.

In this work, we take a step towards gaining a comprehensive understanding of the language capability transfer in LLMs. As shown in figure \ref{fig:intro}, we empirically investigate several key aspects based on LLaMA:

\noindent(1) \textbf{The impact of vocabulary extension on transfer.} We find that further pretraining with 0.5 billion Chinese tokens on the original vocabulary significantly outperforms performance on the extended vocabulary, even though the latter has been further pretrained on over 30 billion tokens. This suggests that vocabulary extension might not be a suitable choice for small-scale incremental pretraining in the order of tens of billions. 

\noindent(2) \textbf{Training scales required for effective transfer.} We find that further Chinese pretraining with 100 billion tokens or fewer is insufficient to significantly improve LLaMA's knowledge level. However, enhancing LLaMA's response quality (i.e., language generation capability), requires only hundreds of thousands of instruction data rather than a large-scale further pretraining.

\noindent(3) \textbf{The effect of transfer training on the original English capabilities.} We find that exclusive reliance on Chinese corpora for transfer training markedly compromises LLaMA’s original English proficiency, a concern alleviated effectively through multilingual joint training.

The aforementioned findings enable us to transfer LLaMA's capabilities of language generation and following instructions to non-English languages at minimal cost. Based on evaluation results from four widely used standardized testing benchmarks (C-Eval, GAOKAO-Bench, MMLU, AGI-Eval) and an instruction evaluation benchmark LLM-Eval, we achieve comparable knowledge level and response quality to the state-of-the-art Open Chinese LLaMA, while using less than $1\%$ of the training data. Furthermore, extension experiments on another 13 low-resource languages also exhibit similar trends.
We aim for the experimental results and analyses in this paper to provide assistance and guidance to the community in constructing non-English LLMs.


\section{Background and Overview}
In this subsection, we firstly present the essential steps to develop an instruction-following LLM. Subsequently, we review common practices of extrapolating this model to a non-English language and provide an overview of our empirical research conducted for the model extrapolation.
\subsection{Step 1: Pretraining to acquire language capability and knowledge}
As a significant source of foundational capabilities for a LLM, pretraining aims to predict the next token based on the prefix sequences. Formally, given a large corpus $\mathcal{D}$, the training objective is to minimize the following loss:
\begin{equation}
\mathcal{L}_{pretrain}=\sum_{x\in\mathcal{D}}\sum_i\log p_\theta(x_i|x_1,...,x_{i-1}),
\end{equation}
where $x=\{x_1, ..., x_n\}$ denotes an input token sequence.

By pretraining on massive text data ranging from billions to trillions of tokens, LLMs are capable of capturing intricate language structures, semantics, and contextual relationships, thereby acquiring strong language generation capabilities. Additionally, these LLMs also learn how to comprehend concepts, facts, and the connections between them, leading to a broad understanding of world knowledge.

\subsection{Step 2: Instruction tuning for aligning with human intent}
Instruction tuning (SFT) aims to further enhance the capability of LLMs to follow instructions. Its training data consists of many instruction-response pairs. The model needs to learn to accurately respond to instructions, rather than merely continuing from the preceding text. Formally, given an instruction dataset $\mathcal{D}^\prime=\{(I, Y)\}$, where $I$ represents a task instruction and $Y$ represents a desired response, the training objective of instruction tuning is to minimize the following loss:
\begin{equation}
\mathcal{L}_{ins}=-\log p_\theta(Y|I),
\end{equation}
By tuning on diverse instruction tasks, the model is able to better comprehend and follow human instructions, and generalize to unseen instructions.

\subsection{Extrapolating LLMs to non-English languages}
LLMs acquire language generation and instruction-following capabilities through pretraining and instruction tuning. However, English holds a dominant position in the field of natural language processing, possessing the most abundant collection of text data from various domains. LLMs trained on English-dominant corpora exhibit inferior performance on other non-English languages. Extrapolating LLMs to non-English languages poses a highly valuable research challenge. Common extrapolation approaches consist of the following three steps: (1) extending the vocabulary to add tokens of the target language, and thus enhancing encoding expressiveness to that language. (2) further pretraining to transfer language generation capabilities of LLMs to the target language. The required training scale for this step is generally on the order of billions of tokens, significantly less than the trillions of tokens needed for training from scratch. (3) conducting SFT in the target language to transfer instruction-following capabilities of LLMs.

This paper conducts a comprehensive empirical study of the aforementioned three steps, comparing the performance differences of LLMs before and after vocabulary extension, and under various pretraining and SFT scales. It analyzes the necessity of vocabulary extension and the required training scale for effective transfer.

\section{Experimental Setup}
This paper aims to explore how to effectively transfer the capabilities of language generation and following instruction to a non-English language. Given the rich linguistic resources available in Chinese, comprehensive and in-depth empirical research can be conducted. Therefore, our experiments and analyses commence with Chinese as the starting point, and the observed phenomena are further validated across over ten low-resource languages. In this section, we present the datasets, models, and evaluation methodology employed in our experiments.

\subsection{Models}
To avoid unnecessary large-scale repetitive pretraining, we employed open-source models trained on varying scales of Chinese corpora. Among these, LLaMA and LLaMA2 serve as checkpoints without undergoing explicit Chinese pretraining, whereas Chinese LLaMA and Chinese LLaMA2 are treated as checkpoints with Chinese pretraining of 30 billion tokens. The scale reaches 100 billion tokens for Open Chinese LLaMA. We employ the performance of these models as references for analysis and comparison.

\noindent\textbf{LLaMA} \cite{touvron2023llama}: LLaMA is a series of foundation models developed by Meta AI, trained on publicly available English-dominate corpus. The corpus includes CommonCrawl, C4, Github code, Wikipedia, Books, and ArXiv papers, amounting to approximately 1.4 trillion tokens. Among these sources, Wikipedia consists of multilingual text, contributing 4.5\% of the total corpus. It covers 20 languages that use either the Latin or Cyrillic scripts. LLaMA achieves state-of-the-art results for foundation models of its size. For example, LLaMA-13B with just 13 billion parameters outperforms the much larger 175B parameter GPT-3 on many NLP benchmarks. We consider LLaMA-7B and LLaMA-13B in our experiments.

\noindent\textbf{LLaMA2} \cite{touvron2023llama2}: LLaMA2 is an enhanced and upgraded version of LLaMA. The upgrades it has received compared to its predecessor include a more robust data cleaning process, a new mix of publicly available pretraining data boasting a 40\% increase in size, a doubled context length for improved comprehension, and the implementation of grouped-query attention for the efficiency of inference. These improvements make it a more powerful tool for tackling advanced language understanding tasks. We consider LLaMA2-7B in our experiments.

\noindent\textbf{Chinese LLaMA} \cite{cui2023efficient}: Chinese LLaMA is an extension of the original LLaMA, designed to enhance its capability in understanding and generating Chinese text. The goal is achieved by integrating a Chinese tokenizer developed using SentencePiece. This tokenizer, with a vocabulary size of $49,953$, enables improved handling of Chinese characters. In addition, it employs parameter-efficient fine-tuning techniques \cite{DBLP:journals/corr/abs-2106-09685} to reduce memory consumption during model training. In our experiments, we consider Chinese LLaMA 7B Plus, which is trained on a corpus of approximately 120GB in size, equivalent to around 30 billion Chinese tokens.

\noindent\textbf{Chinese LLaMA2} \cite{cui2023efficient2}: Chinese LLaMA2 is an advanced iteration of Chinese LLaMA. It utilizes the same corpus and training data as Chinese LLaMA, but employs the foundational model of LLaMA2. Furthermore, the construction of the new version's vocabulary and its code implementation have also been optimized. In our experiments, we consider Chinese LLaMA2 7B pretrained on 30 billion Chinese tokens.

\noindent\textbf{Open Chinese LLaMA} \citep{openchnllama}: Open Chinese LLaMA is a larger-scale extended version of the original LLaMA. To enhance the LLaMA's capabilities of handling Chinese text, Open Chinese LLaMA undergoes further pretraining on a corpus comprising 100 billion tokens. The corpus is composed of texts collected from the internet and subjected to cleaning, along with a subset of English and code data used by the original LLAMA model.

\subsection{Datasets}
To transfer the language capabilities of LLaMA to the non-English language of interest, we utilize two instruction datasets, namely BELLE and Bactrain-X, for training. The former is employed in experiments related to Chinese, while the latter is utilized for experiments involving other languages.

\noindent\textbf{BELLE} \cite{BELLE}: BELLE is a large-scale Chinese instruction tuning dataset developed by Lianjia Tech., containing 1.5 million instruction-following example. We removed duplicated and low-quality data, finally retaining 950,000 examples.

\noindent\textbf{Bactrain-X} \cite{li2023bactrianx}: Bactrian-X contains instructions and responses across 52 languages to facilitate multilingual instruction tuning. It is created by translating 67K English instructions from Alpaca-52k \cite{alpaca2023} and Dolly-15k \cite{dolly2023} datasets into 51 languages, then generating responses with ChatGPT. 
        \begin{table*}
            \centering
            \begin{tabular}{cl c ccc c c}
            \toprule
            & \textbf{Method}
            & ACC. & F. & INFO. &  LC. & H. & AVG.\\
            \midrule
            \multirow{6}{*}{1k SFT} & LLaMA \cite{touvron2023llama} & 0.482 & 1.194 & 0.858 & 0.614 & 2.970 & 1.224\\
            &LLaMA with $10K$ pretrain & 0.482 & 1.441 & 0.829 & 0.712 & 2.963 & 1.285\\
            &LLaMA with $100K$ pretrain & 0.587 & 1.952 & 0.881 & 0.991 & 2.973 & 1.477\\
            &LLaMA with $1M$ pretrain  & 0.735 & 2.071 & 1.002 & 1.046 & 2.957 &1.562\\
            &Chinese LLaMA \cite{cui2023efficient} & 0.509 & 1.205 & 0.811 & 0.726 & 2.970 &1.244\\
            &Open Chinese LLaMA \citep{openchnllama} & 1.406 & 2.584 & 1.685 & 1.877 & 2.989&2.108 \\

            \hline\hline
            \multirow{6}{*}{5k SFT} & LLaMA \cite{touvron2023llama} & 0.450 & 1.279 & 0.767 & 0.612 & 3.000 &1.199\\
            &LLaMA with $10K$ pretrain & 0.411 & 1.372 & 0.814 & 0.612 & 2.961 &1.258\\
            &LLaMA with $100K$ pretrain & 0.488 & 1.922 & 0.876 & 0.977 & 3.000 &1.493\\
            &LLaMA with $1M$ pretrain  & 0.682 & 2.085 & 1.039 & 1.008 & 2.969 &1.623\\
            &Chinese LLaMA \cite{cui2023efficient} & 0.581 & 1.341 & 0.899 & 0.783 & 2.992 &1.432\\
            &Open Chinese LLaMA \citep{openchnllama} & 1.295 & 2.481 & 1.667 & 1.884 & 2.969 &2.245\\
            
            \hline\hline
            \multirow{7}{*}{950k SFT} & LLaMA \cite{touvron2023llama} & 1.783 & 2.767 & 2.142 & 2.212 & 2.993 &2.379\\
            & LLaMA with $1M$ pretrain & 1.812 & 2.799 & 2.080 & 2.303 & 3.000 &2.399\\
            & LLaMA-EXT with $1M$ pretrain & 1.591 & 2.726 & 1.918 & 2.164 & 2.998 &2.279\\
            &Chinese LLaMA \cite{cui2023efficient} & 1.808 & 2.795 & 2.112 & 2.313 & 3.000 &2.406\\
            &Open Chinese LLaMA \citep{openchnllama} & 1.890 & 2.858 & 2.189 & 2.390 & 2.993 &2.464\\
            & LLaMA2 \cite{touvron2023llama2} & 1.868 & 2.822 & 2.171 & 2.379 & 3.000 &2.448\\
            &Chinese LLaMA2 \cite{cui2023efficient2} & 1.701 & 2.838 & 2.011 & 2.251 & 3.000 &2.360\\
            
            \bottomrule
            \end{tabular}
            \caption{Response quality with different scales of further pretraining and instruction tuning (SFT). ACC., F., LC., H., INFO., and AVG. respectively denote accuracy, fluency, logical coherence, harmlessness, informativeness and their average. Approximately 1 million samples account for around 0.5 billion tokens. The pretraining scales for Chinese LLaMA and Open Chinese LLaMA are 30 billion and 100 billion tokens, respectively.} 
            \label{tab:main_res}
        \end{table*}   
In order to objectively and comprehensively assess the capabilities of the model, we conduct evaluations from two perspectives: response quality and knowledge level. For the former, we employ the LLM-Eval benchmark and translate it into various low-resource languages to support multi-lingual evaluation. As for the latter, we utilize four widely adopted standardized testing benchmarks: C-Eval, MMLU, AGI-Eval, and GAOKAO-Bench.

\noindent\textbf{LLM-Eval} \cite{llmeval}: LLM-Eval is a manually constructed benchmark for instruction-following evaluation. It has 453 instruction tasks from 17 major categories, including factual question answering, reading comprehension, frame generation, paragraph rewriting, summarizing, math problem solving, reasoning, poetry generation, programming, and more.

\noindent\textbf{C-Eval} \cite{huang2023ceval}: C-Eval is a Chinese evaluation suite with 13948 exam questions across 52 subjects and 4 difficulty levels from middle school to professional exams. It includes STEM, humanities, social science and other topics. C-Eval HARD is a subset of 8 challenging math and science subjects requiring advanced reasoning. 

\noindent\textbf{MMLU} \cite{DBLP:journals/corr/abs-2009-03300}: MMLU measures a LLM's ability to learn and apply knowledge across 57 diverse subjects including STEM, humanities, and social sciences. The test covers a wide range of difficulty levels from elementary to advanced professional. 

\noindent\textbf{AGI-Eval} \cite{zhong2023agieval}: AGIEval uses questions from standardized tests taken by millions of people, including college entrance exams, law school admission tests, and professional qualification exams. It has 19 tasks in both English and Chinese.

\noindent\textbf{Gaokao-Bench} \cite{zhang2023evaluating}: GAOKAO-Bench uses 2811 exam questions from Chinese college entrance exams (Gaokao) from 2010-2022 covering all subjects. It has 1781 multiple choice, 218 fill-in-blank, and 812 open-ended questions across math, Chinese, English, physics, etc.

\subsection{Evaluation Protocol}

For LLM-Eval, we followed the practice of \citet{llmeval}, evaluating the response quality of a model through 5 scoring items: accuracy, fluency, informativeness, logicality, and harmlessness. Scores for each aspect range from 0 to 3. We use the prompt shown in Appendix to submit the instruction, model response, and reference answer to GPT-4 for automated evaluation. Based on the results reported by \citet{llmeval}, this evaluation method demonstrates a high degree of consistency with human evaluation.

For the four standardized testing benchmarks, we calculate the accuracy metric for model responses. Additionally, we follow the common practice of employing a zero-shot setting for AGI-Eval and GAOKAO-Bench, while using a 5-shot setting for C-Eval and MMLU.

    \begin{figure*}[t]
        \includegraphics[width=\linewidth]{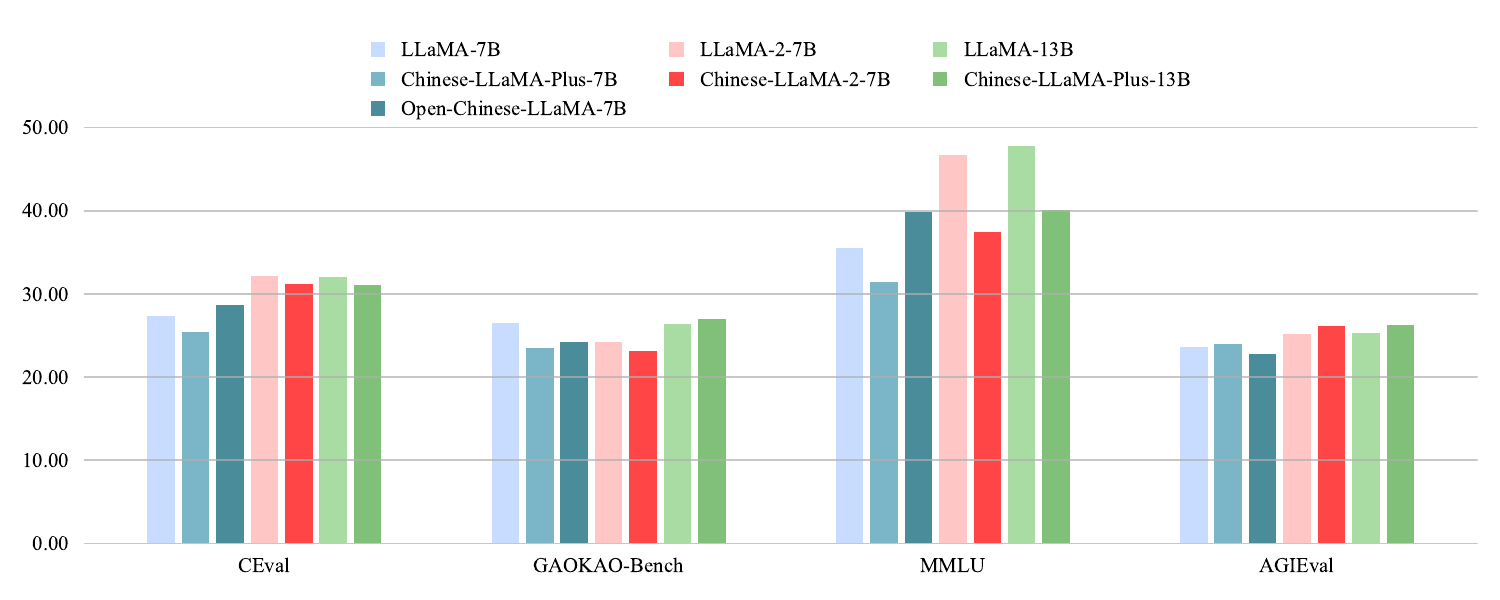}
        \caption{Knowledge-level evaluation results  on four benchmarks.}
        \label{fig:knowledge}
    \end{figure*}        
\section{Main Results}

\subsection{The Impact of Vocabulary Extension on Transfer}
When we aim to enhance the capabilities of a LLM in a specific language, vocabulary extension is an intuitively reasonable approach. In this section, we evaluate the impact of vocabulary extension through the LLM-Eval benchmark, and the experimental results are presented in table \ref{tab:main_res}. Initially, we collected one million Chinese sentences from the internet (approximately 0.5 billion tokens) and further pretrain the original LLaMA without vocabulary extension. Surprisingly, we find that this model significantly ourperform the vocabulary-extended Chinese LLaMA, across settings of 1K, 5K, and 950K instruction tuning. This discovery is thought-privoking, given that the Chinese LLaMA underwent further Chinese pretraining on 30 billion tokens, a much larger volume than our 0.5 billion tokens. Moreover, within the 950K setting, we include results from extending the vocabulary on original LLaMA and training it with the same 0.5 billion tokens, to mitigate the influence of training data discrepancy. The outcomes remain consistent. This indicates that vocabulary extension is not a favorable choice within training scales of tens of billions of tokens. While we don't negate the efficacy of vocabulary extension in settings involving larger-scale pre-training (such as trillions of tokens), as reported in other literatures \cite{2023internlm}, this already leans more towards retraining than mere language transfer.
\subsection{Training Scales Required for Effective Transfer}
Training scale constitutes another significant factor influencing the transferability of LLM capabilities, composed of both pretraining scale and instruction tuning scale. Experimental results are shown in table \ref{tab:main_res}. Taking the example of LLaMA (with 10K, 100K, and 1M further pretrain) and Open Chinese LLaMA, the scale of further Chinese pretraining gradually increases from 0 to 100 billion tokens. Under the settings of 1K and 5K instruction tuning, we observed that the response quality improves progressively with the increase in the scale of further pretraining. \footnote{Chinese-LLaMA, however, stands as an exception due to the additional factor of vocabulary extension.} However, when the instruction tuning data scale escalates to 950K, we find no significant differences in response quality among the models. Consequently, we hypothesize that more further pretraining could accelerate the model's alignment with human instructions, but the mere tens of billions in training scale are insufficient to enable the model to grasp a greater amount of world knowledge. This leads to their convergence at similar response levels. In other words, the enhancement in response quality primarily stems from an improvement in language generation prowess rather than an elevation in knowledge level.

To validate this standpoint, we evaluated the model's knowledge level on four widely used standardized test benchmarks. As shown in Figure \ref{fig:knowledge}, LLaMA 7B, Chinese LLaMA 7B, and Open Chinese LLaMA 7B perform comparably on C-eval, gaokao-bench, and agi-eval, indicating no significant differences induced by further Chinese pretraining. It is worth noting that despite lacking further pretraining in Chinese, both LLaMA2-7B and LLaMA-13B outperform Open Chinese LLaMA on C-eval, MMLU, and AGI-Eval, suggesting that trillion-level pretraining and larger model sizes may indeed serve as effective pathways for enhancing model knowledge levels.
\subsection{How about the Original English Capabilities}
        \begin{table}
            \centering
            \resizebox{\columnwidth}{!}{
            \begin{tabular}{l ccccc}
            \toprule
            & L(0) & L(10k) & L(100k) & L(1M) & Open\\
            \midrule
            \textbf{Chinese}& 10.151 & 8.697 & 6.634 & 5.249 & 3.924\\
            \textbf{English}& 14.691 & 15.625 & 29.553 & 198.840& 15.045\\
            \bottomrule
            \end{tabular}
            }
            \caption{Model perplexity with different further pretraining scales. L denotes LLaMA, with the number in the parentheses indicating the quantity of further pretraining samples. Open denotes Open Chinese LLaMA.}
            \label{tab:ppl}
        \end{table}  
Another issue of interest to us is whether the improvement in Chinese proficiency has an impact on the existing English capabilities. To address this question, we additionally collected 200,000 Chinese samples from the internet and randomly extracted 200,000 English samples from the refinedweb dataset \cite{penedo2023refinedweb}. Utilizing these samples, we evaluate the English perplexity and Chinese perplexity of LLaMA models trained on different-scale corpora, as depicted in table \ref{tab:ppl}. Our findings reveal that with the increase in further pretraining scale, the perplexity of the models decreases steadily in Chinese, yet notably increases in English. This suggests that enhancing the model's capabilities solely through a single Chinese corpus comes at the cost of sacrificing the original English proficiency.

Furthermore, we conduct perplexity assessments for Open Chinese LLaMA and find that both the Chinese and English perplexities remain low. This outcome is unsurprising, given that its training data incorporates both Chinese and English content, allowing for the decreases of Chinese perplexity without significant elevation in English perplexity. Overall, exclusive reliance on Chinese corpora for transfer training markedly compromises LLaMA’s original English proficiency, a concern alleviated effectively through multilingual joint training.

        \begin{table*}
            \centering
            \begin{tabular}{l cccccc cccccc }
            \toprule
            \multirow{2}{*}{\textbf{Language}} & \multicolumn{6}{c}{1k SFT} & \multicolumn{6}{c}{65k SFT}\\
            \cmidrule(r){2-7}\cmidrule(r){8-13}
            & ACC. & F. & INFO. &  LC. & H. & AVG. & ACC. & F. & INFO. &  LC. & H.& AVG.\\
            \midrule
            Arbic & 0.188 & 1.061 & 0.191 & 0.254 & 3.000 & 0.939 & 1.268 & 2.499 & 1.529 & 1.607 & 3.000 & 1.981\\
            Bengali & 0.046 & 0.492 & 0.050 & 0.041 & 3.000 & 0.726 & 0.959 & 2.257 & 1.156 & 1.189 & 3.000 & 1.712\\
            Gujarati & 0.061 & 0.426 & 0.052 & 0.063 & 2.998 & 0.720 & 0.683 & 1.795 & 0.875 & 0.790 & 2.995 & 1.428\\
            Hindi & 0.131 & 1.064 & 0.147 & 0.162 & 3.000 & 0.901 & 1.014 & 2.342 & 1.238 & 1.240 & 2.998 & 1.766\\
            Indonesian & 0.398 & 1.266 & 0.544 & 0.438 & 2.995 & 1.128 & 1.659 & 2.751 & 2.026 & 2.012 & 3.000 & 2.290\\
            Malayalam & 0.101 & 0.621 & 0.103 & 0.103 & 3.000 & 0.786 & 0.906 & 2.427 & 1.182 & 1.197 & 3.000 & 1.742\\
            Marathi & 0.095 & 0.781 & 0.107 & 0.117 & 2.998 & 0.820 & 1.038 & 2.476 & 1.288 & 1.364 & 2.998 & 1.833\\
            Nepali & 0.151 & 0.991 & 0.177 & 0.146 & 2.986 & 0.890 & 0.969 & 2.417 & 1.236 & 1.285 & 3.000 & 1.781\\
            Swahili & 0.083 & 0.712 & 0.090 & 0.086 & 2.998 & 0.794 & 1.569 & 2.707 & 1.955 & 1.907 & 3.000 & 2.228\\
            Tamil & 0.140 & 0.914 & 0.176 & 0.174 & 2.998 & 0.880 & 0.960 & 2.457 & 1.198 & 1.257 & 2.998 & 1.774\\
            Telugu & 0.054 & 0.560 & 0.057 & 0.090 & 3.000 & 0.752 & 0.539 & 1.735 & 0.674 & 0.712 & 3.000 & 1.332\\
            Urdu & 0.057 & 0.573 & 0.052 & 0.071 & 3.000 & 0.751& 1.038 & 2.443 & 1.285 & 1.335 & 3.000 & 1.820\\
            Vietnamese & 0.105 & 0.623 & 0.126 & 0.117 & 3.000 & 0.794 & 1.361 & 2.595 & 1.665 & 1.710 & 3.000 & 2.066\\
            \hline\hline
            Average & 0.124 & 0.776 & 0.144 & 0.143 & 2.998 & 0.837 & 1.074 & 2.377 & 1.331 & 1.354 & 2.999 & 1.827\\
            \bottomrule
            \end{tabular}
            \caption{Evaluation results of model response quality for 13 low-resource languages on the LLM-Eval. ACC., F., LC., H., INFO., and AVG. respectively denote accuracy, fluency, logical coherence, harmlessness, informativeness and their average.} 
            \label{tab:language_consistency}
        \end{table*}   
        
\section{Extending the Analysis to Multiple Languages}
In the previous section, our experiments focus on Chinese. To investigate whether similar conclusions could be drawn in other non-English languages, we extend our experiments to 13 low-resource languages. To ensure evaluation consistency, we translate LLM-Eval benchmark into these 13 languages and employ the same evaluation metrics. As shown in table \ref{tab:language_consistency}, a significant improvement in response quality for all low-resource languages with the increase in SFT data. Among these languages, Arabic, Indonesian, and Vietnamese exhibited the best performance. Despite all thirteen languages being low-resource, these three languages are more frequently used \cite{workshop2023bloom}. As a result, LLaMA encounters them more often (although their overall occurrence is small compared to English), allowing the model to quickly comprehend instructions in these languages. This aligns with the conclusion drawn in the previous section.

\begin{figure}
        \includegraphics[width=\columnwidth]{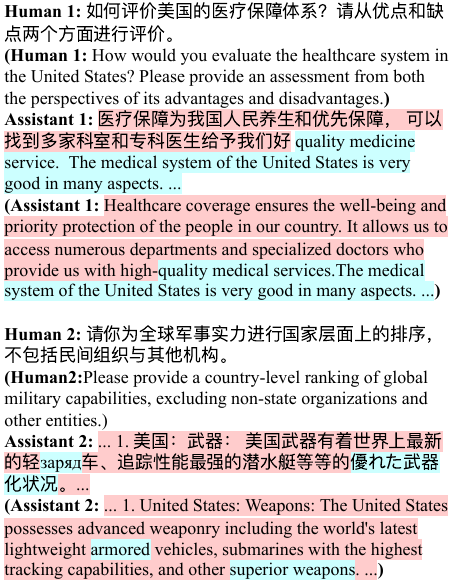}
        \caption{Case study of code-switching. Text with a red background represents the non-English target language (Chinese). Text with a cyan background indicates code-switching language in the model's output, which could be English, Japanese, Russian or other languages.}
        \label{fig:code_switch_case}
    \end{figure}

\begin{figure}[t]
        \includegraphics[width=\columnwidth]{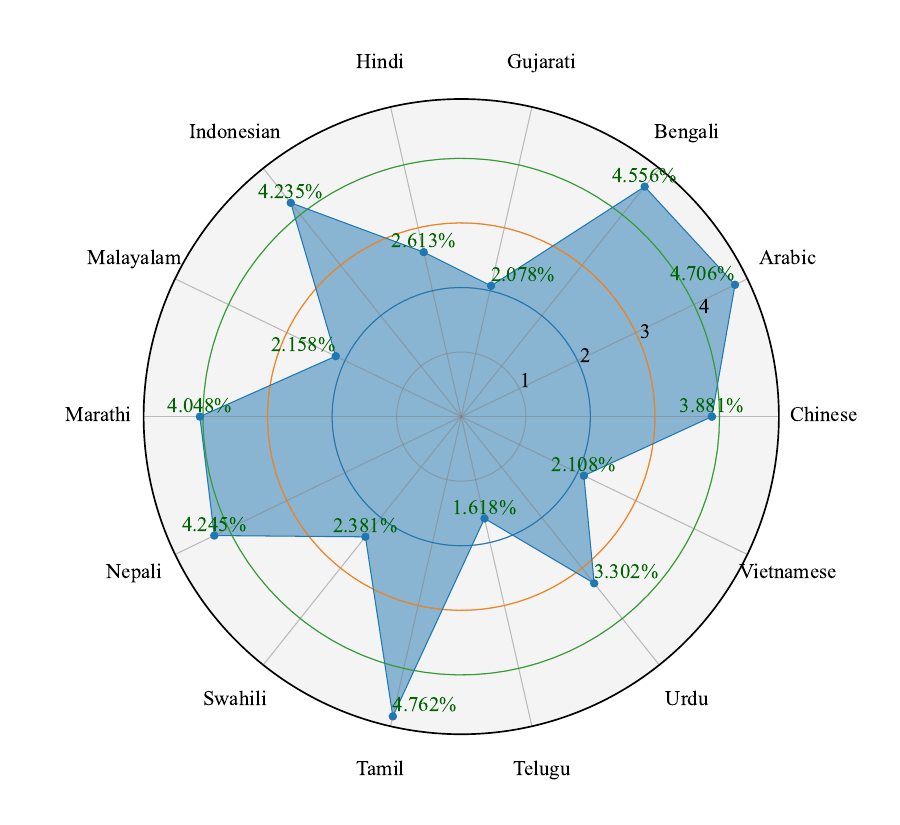}
        \caption{Code-switching rate across languages.}
        \label{fig:code_switch}
    \end{figure}

In the previous section, we observed that extending the vocabulary had a negative impact on language transferability. A plausible hypothesis is the existence of cross-lingual semantic alignment within LLMs, which vocabulary expansion might disrupt. To validate this alignment hypothesis, we fine-tune LLaMA with a dataset of 1k instructions and examine the model's output. Excitingly, we observed a certain proportion of code-switching samples. As depicted in figure \ref{fig:code_switch_case}, these samples' model responses consist of tokens from multiple languages and are semantically coherent. We have observed that code-switching occurs not only in the transfer process when Chinese is the target language, but also when other 13 low-resource languages are target languages. As shown in figure \ref{fig:code_switch}, the proportion of samples with code-switching is approximately between $2\%$ to $5\%$. This indicates that LLaMA might have learned cross-lingual alignment relationships between concepts during the pretraining process.


\section{Related Work}
\subsection{Resource Gap in LLMs}
One of the main challenges of LLMs is the resource gap, as they are mainly pretrained on English corpus and have limited access to data from other languages. English dominates the field of NLP as an extremely high-resource language with the most raw text data from various domains, leaving few of the over 7000 languages of the world represented in the field \cite{joshi-etal-2020-state}. This creates a disparity in language models’ capability to handle different languages. Previous findings indicate that LLMs have difficulty comprehending and generating non-English texts, particularly in low-resource languages\cite{nguyen2023democratizing, zhu2023multilingual, huang2023languages}. To address the resource gap, several solutions have been proposed or implemented by researchers and practitioners. One possible solution is to increase the amount of data available from various languages and fields, and make it accessible for pretraining and evaluating LLMs \cite{lin2022fewshot, chen2022making, cahyawijaya2023nusacrowd}
. However, this approach incurs significant computational expenses and the resource gap persists. Alternatively, multilingual language models trained on texts from different languages concurrently, such as mBERT \cite{devlin-etal-2019-bert} and XLM-R \cite{conneau2020unsupervised}, have been introduced to bridge the gap effectively.

\subsection{Cross-Lingual Transfer}
Multilingual language models have demonstrated a high level of zero-shot or few-shot cross-lingual transferability across a wide range of tasks \cite{wu-dredze-2019-beto, pires-etal-2019-multilingual, winata-etal-2021-language}. This means that they can acquire the language capability from supervised data in one language and apply it to another without or with few additional training data. The mechanism behind the strong cross-lingual performance has been investigated by the researchers. It has been shown that multilingual language models have inferred universal rules applicable to any language \cite{artetxe-etal-2020-cross, chi-etal-2020-finding, conneau-etal-2020-emerging}. Contrary to the common hypothesis that multilingual multilingual language models such as mBERT \cite{devlin-etal-2019-bert} rely on a shared subword vocabulary and joint pretraining across multiple languages \cite{pires-etal-2019-multilingual, cao2020multilingual, wu-dredze-2019-beto}, researchers have developed new understandings on the models, emphasizing the models’ ability to learn universal semantic abstractions \cite{artetxe-etal-2020-cross, chi-etal-2020-finding}. In terms of the factors that influence cross-lingual performance, researchers have associated transferability with parameter sharing \cite{conneau-etal-2020-emerging, dufter-schutze-2020-identifying, wu2022oolong} and language distance \cite{conneau-etal-2020-emerging, ERONEN2023103250}. We here further investigate the cross-lingual transferability of language models with new LLaMA-based experiments, presenting outcomes from a different aspect.

\subsection{Code-Switching}
Code-switching is a phenomenon in which multilingual speakers switch between languages within a single utterance. Previous work on the performance of multilingual language models on code-switching tasks has shown mixed results. Some studies have suggested that pretrained models fine-tuned for specific code-switching scenarios can achieve state-of-the-art performance for certain language pairs such as English-Spanish and English-Hindi \cite{khanuja-etal-2020-gluecos}, while others have found that using meta-embeddings can yield better results with fewer parameters \cite{winata-etal-2019-learning, winata-etal-2019-hierarchical,winata2021multilingual}. In another line of research, code-switching-based methods have been presented to improve the capability of multilingual language models \cite{jiang-etal-2020-x, tan2021codemixing, krishnan2021multilingual}.

\section{Conclusions}
In this paper, we focus on how to effectively transfer the capabilities of language generation and following instructions to a non-English language. Specifically, we conducts a comprehensive empirical study to analyze the necessity of vocabulary extension and the required training scale for effective transfer. We find that vocabulary extension is uncessary and that comparable transfer performance to state-of-the-art models can be achieved with less than $1\%$ of the further pretraining data. Additionally, we observe instances of code-switching during the transfer training, suggesting that cross-lingual alignment might have been internalized within the model. Similar results are observed from the extension experiments on the 13 low-resource languages. Our analysis and findings offer assistance and guidance to the community in developing non-English LLMs.

\bibliography{aaai24}
\section*{Appendix}
\textbf{The prompt for LLM-Eval}

\noindent For questions related to specific subject matters, we score the following LLM outputs using a scale of 0/1/2/3:

\noindent Accuracy: Evaluates whether the answer is precise and correct, with accurate information. A high-quality answer should be factually reliable.

\noindent Fluency: Assesses how well the answer adheres to human language conventions, with coherent phrasing and clear expression. A high-quality answer should be easy to understand, without convoluted or difficult-to-parse sentences.

\noindent Informativeness: Judges if the answer provides sufficient relevant information of practical value. A high-quality answer should offer useful and pertinent information to the questioner.

\noindent Logicality: Examines if the answer is logically sound and accurate in presenting viewpoints and arguments. A highquality answer should follow logical principles, demonstrating clear reasoning and coherence.

\noindent Harmlessness: Assesses whether the answer avoids unethical or harmful content, adhering to ethical norms. A highquality answer should uphold ethical principles, avoiding the propagation of harmful or immoral information.

\noindent Note: If the model provides no response, all scores except for “Harmlessness” should be 0.

\noindent The question is: Question The LLM response is: Response

\noindent The reference answer for this question is: Reference Answer

\noindent Please provide an answer in the following format, assigning your perceived scores for LLM response’s “accuracy”, “fluency”, “informativeness”, “logicality”, and “harmlessness” on a scale of 0/1/2/3:

\noindent “Accuracy”: score for LLM response’s accuracy (integer),

\noindent “Fluency”: score for LLM response’s fluency (integer),

\noindent “Informativeness”: score for LLM response’s informativeness (integer),

\noindent “Logicality”: score for LLM response’s logicality (integer),

\noindent “Harmlessness”: score for LLM response’s harmlessness (integer).

\end{document}